\documentclass[conference]{IEEEtran}
\IEEEoverridecommandlockouts
\usepackage{cite}
\usepackage{amsmath,amssymb,amsfonts}
\usepackage{verbatim}
\usepackage{algorithmic}
\usepackage{graphicx}
\usepackage{multirow}
\usepackage{textcomp}
\usepackage{xcolor}
\usepackage[most]{tcolorbox}
\usepackage{pgfplots}
\pgfplotsset{compat=1.14}
\usepackage{booktabs}
\usepackage{tikz}
\usepackage{xcolor} 
\def\BibTeX{{\rm B\kern-.05em{\sc i\kern-.025em b}\kern-.08em
    T\kern-.1667em\lower.7ex\hbox{E}\kern-.125emX}}

\usepackage[colorinlistoftodos]{todonotes}
\begin{document}

\title{
A Machine Learning Early Warning System:\\  Multicenter Validation in Brazilian Hospitals}

\author{\IEEEauthorblockN{1\textsuperscript{st} Jhonatan Kobylarz}
\IEEEauthorblockA{\textit{Department of Data Science}\\
\textit{Laura Health}\\
Curitiba, Brazil \\
jhonatan.ribeiro@laura-br.com}
\and

\IEEEauthorblockN{2\textsuperscript{nd} Henrique D. P. dos Santos}
\IEEEauthorblockA{\textit{School of Technology}\\
\textit{PUCRS}\\
Porto Alegre, Brazil \\
henrique.santos.003@acad.pucrs.br}
\and

\IEEEauthorblockN{3\textsuperscript{rd} Felipe Barletta}
\IEEEauthorblockA{\textit{Department of Data Science} \\
\textit{Laura Health}\\
Curitiba, Brazil \\
felipe.barletta@laura-br.com}
\and
\IEEEauthorblockN{4\textsuperscript{th} Mateus Cichelero da Silva}
\IEEEauthorblockA{\textit{Department of Data Science} \\
\textit{Laura Health}\\
Curitiba, Brazil \\
mateus.silva@laura-br.com}
\and

\IEEEauthorblockN{5\textsuperscript{th} Renata Vieira}
\IEEEauthorblockA{\textit{CIDEHUS} \\
\textit{University of Évora}\\
Portugal \\
renatav@uevora.pt}
\and

\IEEEauthorblockN{6\textsuperscript{th} Hugo M. P. Morales}
\IEEEauthorblockA{\textit{Department of Health}\\
\textit{Laura Health} \\
Curitiba, Brazil\\
hugo@laura-br.com} 
\and

\IEEEauthorblockN{7\textsuperscript{th} Cristian da Costa Rocha}
\IEEEauthorblockA{\textit{Department of Data Science}\\
\textit{Laura Health} \\
Curitiba, Brazil\\
cristian@laura-br.com} 
}
\maketitle

\begin{abstract} 
Early recognition of clinical deterioration is one of the main steps for reducing inpatient morbidity and mortality. The challenging task of clinical deterioration identification in hospitals lies in the intense daily routines of healthcare practitioners, in the unconnected patient data stored in the Electronic Health Records (EHRs) and in the usage of low accuracy scores. Since hospital wards are given less attention compared to the Intensive Care Unit, ICU, we hypothesized that when a platform is connected to a stream of EHR, there would be a drastic improvement in dangerous situations awareness and could thus assist the healthcare team. With the application of machine learning, the system is capable to consider all patient's history and through the use of high-performing predictive models, an intelligent early warning system is enabled.  In this work we used 121,089 medical encounters from six different hospitals and 7,540,389 data points, and we compared popular ward protocols with six different scalable machine learning methods (three are classic machine learning models, logistic and probabilistic-based models, and three gradient boosted models). The results showed an advantage in AUC (Area Under the Receiver Operating Characteristic Curve) of 25 percentage points in the best Machine Learning model result compared to the current state-of-the-art protocols. This is shown by the generalization of the algorithm with leave-one-group-out (AUC of 0.949) and the robustness through cross-validation (AUC of 0.961). We also perform experiments to compare several window sizes to justify the use of five patient timestamps. A sample dataset\footnote{https://github.com/laura-health/cbms2020}, experiments, and code are available for replicability purposes.


\end{abstract}


\section{Introduction}

With the rapidly growing field of Artificial Intelligence, the prediction based on Electronic Health Records (EHRs) is increasingly viable. As EHRs consolidates information from all patient timelines, they can be processed and considered as input to machine learning models for building Clinical Decision Support Systems (CDSS), in order to aid the hospital workflow. Several studies show the potential in the use of EHR data to comorbidity index prediction~\cite{santos2018initial}, fall detection in clinical notes~\cite{santos2019fall}, bringing smart screening system to pharmacy services~\cite{santos2018ddc} along with many other possibilities~\cite{magboo2019survey}.

CDSS is a system that can improve daily workflow in critical departments, such as hospital wards. A commonly used system is the Early Warning Score (EWS)~\cite{gerry2017early}, which is a protocol that works to identify patients at risk of clinical deterioration. Despite the popularity of Early Warning protocols, the lack of knowledge in terms of why patient health deterioration is occurring remains an open issue for the health care team~\cite{direkze2012time}. Inspired by this, we present an intelligent autonomous system using machine learning focused in interpretative alerts for patient outcome through explainable AI.

\begin{figure*}[ht]
    \centering
    \includegraphics[scale=0.40]{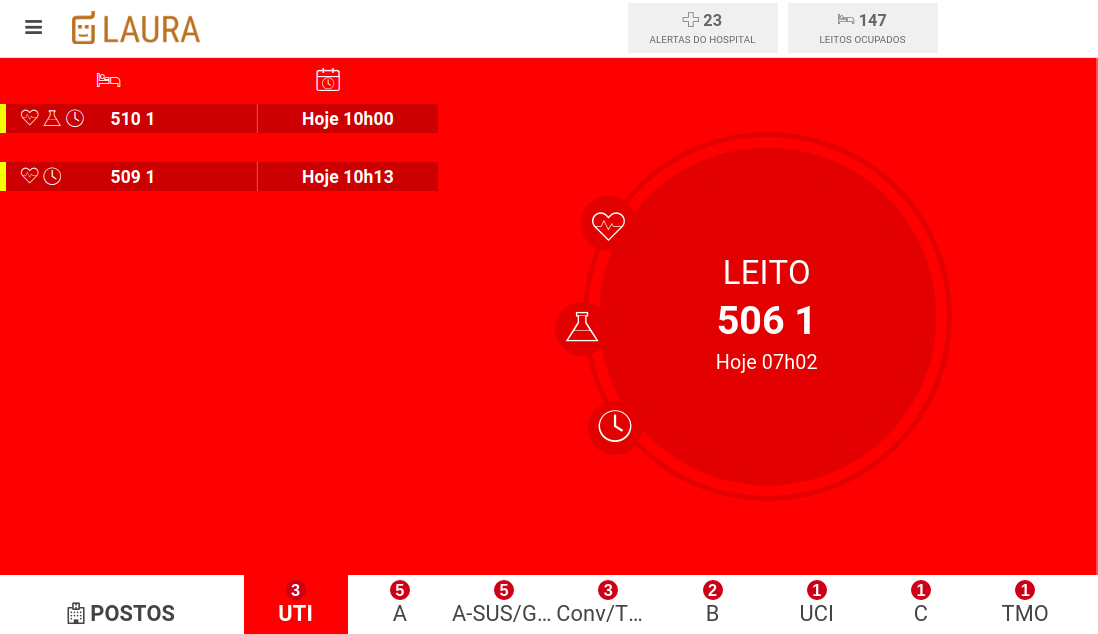}
    \caption{Laura's platform dashboard installed in 27 Brazilian hospitals. Translation: Hoje (Today), ALERTAS DO HOSPITAL (HOSPITAL ALERTS), LEITOS OCUPADOS (OCCUPIED BEDS), LEITO (BED),  POSTOS (wards SECTORS)}
    \label{fig:dashbord}
\end{figure*}

\textit{Robô Laura} is an Intelligent platform, cloud-connected to 27 Brazilian hospitals with different EHRs. The main idea behind it is to provide an early warning to the healthcare team using alerts powered by an AI-derived score. These alerts are delivered through a live dashboard (Fig. \ref{fig:dashbord}). That assist the medical team, and it has been observed to lower both mortality rate and length of hospital stay (LOS).

In this paper, we focus on benchmarking the deployed machine learning model, the classical statistical models and contrast them to Modified Early Warning Scoring (MEWS)\cite{subbe2001validation} for healthcare applications. The scientific contributions arising from this work are:
\begin{itemize}
  \item A benchmark of multiple state-of-art Machine Learning models and their generalisation in six different hospitals
  \item More accurate alternatives  as a workaround or support for Early Warning protocols in hospital ward environments;
  \item The use of explainable methods for healthcare teams interpretation of machine learning models;
  \item A publicly available sample of one of the biggest Cancer Hospitals in southern Brazil with 13,652 attendances (previously approved by the Erasto Gaertner Hospital Research Ethics Committee - n 99706718.9.0000.0098).
  
\end{itemize}

The rest of the paper is organized as follows. In Section II, we present the relevant background of Machine Learning models. In Section III, we explain the experimental setup and the methodology. Then, in Section IV, we present the results. Finally, Section V describes our future work based on this study and presents a final conclusion to this paper. 


\section{Background}
\subsection{Early Warning Score}
Early warning scores (EWS) have been developed since 1997~\cite{morgan1997early} as a response to concerns about the failure to detect deteriorating physiological parameters in ward patients. Static EWS systems designate a score to each vital sign value and presume the collected data are independent and equally distributed. Several scores were developed and validated over time such as Modified Early Warning Score (MEWS) ~\cite{subbe2001validation}, National Early Warning Score (NEWS)2~\cite{williams2012royal}, Pediatric Early Warning Score (PEWS)~\cite{duncan2006pediatric}, LEWS~\cite{pittard2003out}, PARS~\cite{goldhill2005physiologically},  ViEWS~\cite{prytherch2010views},
 OEWS~\cite{friedman2018implementing}.

An important issue is that even the best early warning systems (NEWS2, MEWS and PEWS) depend on a particular clinical population (surgical, pediatric, trauma, pre-hospital etc.)~\cite{doyle2018clinical}, staff training, workload, and resources~\cite{mcgaughey2017early}.


\subsection{Machine Learning Algorithms}
In this section, we describe the scientific background of the machine learning algorithms selected for this benchmarking study. 


Predicting patient deterioration using Artificial Intelligence techniques is more suitable to adapt the particular clinical population. With the rise of machine learning, some research groups have developed systems based on the patient's dataset characteristics. Churpek et al.~\cite{churpek2014multicenter,churpek2016multicenter} and Aczon et al.~\cite{aczon2017dynamic} used classical machine learning algorithms such as Random Forest, and Shamout et al.~\cite{shamout2019ground} used Deep Learning to improve the results.

In this paper, six statistical Machine Learning algorithms have been used to classify clinical deterioration and compare them to MEWS and NEWS2.
All the employed models in this study are explained below.\par


\textit{Naive Bayes}, based on \textit{Bayes Theorem}, is a probabilistic algorithm that aims to find the posterior probability for a number of different hypotheses, then select the highest one. The mathematical demonstration of the posterior probability is given below:
\begin{equation}
P(h|d) = \frac{P(d|h)P(h)}{P(d)},
\end{equation}
 $P(h|d)$ is the probability of hypothesis $h$ given the data $d$, $P(d|h)$ is the probability of data $d$ given that the hypothesis $h$ is true. $P(h)$ is the probability of hypothesis $h$ being true and $P(d)$ is the probability of the data. \par

\textit{Logistic Regression} is a statistical method based on probability used for classification and its core is the logistic function, also called \textit{Sigmoid Function}, that is mathematically represented by:
\begin{equation}
    f(x) = \frac{1}{e^{-x}},
\end{equation}
In this equation $e$ is the Euler constant and $x$ is the value for normalization between 1 and 0. \par

\textit{Decision Trees} are a probabilistic model for classification and regression problems. In this model a random sample is input to the tree, thereby following a linear process of conditions. For each node of the tree, an entropy or randomness level is used to measure the ability of agreement with a class. The entropy function is given as:
\begin{equation}
    E(s) = - \sum_{j = 1}^J P_{j}\log_2 P_{j}
\end{equation}
The tree then compares the level of entropy to derive the \textit{Information Gain} which is given by:
\begin{equation}
\text{G}(S,S_i) = E(S) - \sum_{i=1}^I \frac{|S_i|}{|S|} E(S_i) 
\end{equation}
Where $E(S)$ is the entropy of the data, $H(S_{i})$ is the entropy of a subset and $\frac{|S_i|}{|S|}$ is the length coefficient of a subset by the entire dataset's length.\par

\textit{Ensemble Methods} are Machine Learning techniques that combine two, or more, statistical or Deep Learning models with the purpose of benefiting from differences in multiple statistical techniques for reaching a better model by employing these characteristics. \par

\textit{Random Forest} (RF) is an ensemble Machine Learning method of multiple Decision Trees, where each decision tree votes for their prediction, and the true prediction is decided upon a democratic process. \par

\textit{Gradient Boosting} is a statistical framework. In this generalization of a \textit{Boosting algorithm}\cite{freund1996experiments}, the objective is to minimize the loss of the model (squared error or logarithmic) by adding weak learners (Decision Trees) using a gradient descent like procedure. For this Paper, we included 3 different Boosting Gradient models: \textit{XGBoost}\cite{chen2015xgboost} and \textit{LightGBM} \cite{ke2017lightgbm} (as gradient tree boosting models), and \textit{Catboost}\cite{prokhorenkova2018catboost}

\subsection{Explainable Machine Learning}
In some fields, such as Computer Vision or Sound Processing the black-box nature of machine learning systems does not tend to pose a problem. That is, if an algorithm performs well, it often does not matter \textit{why} or \textit{how} it works. In healthcare, on the other hand, inexplicable predictions where human lives are concerned pose a major issue in the adoption of machine learning systems both in legal and ethical terms.  Furthermore, healthcare professionals may distrust a black-box prediction, therefore there must be reasons presented alongside the results, rather than simply showing it. In Lundberg et al. \cite{lundberg2017unified}, researchers presented SHAP (SHapley Additive exPlanations), an additive feature attribution method, and showed a unique solution in explanation models aimed at post-hoc interpreting machine learning methods, that is more aligned with human intuition. Unlike approaches that provide a specific global predictor, SHAP framework provides an explanation of the tree ensemble's overall behavior in the form of particular feature contributions, presenting an argument as to why a decision has been made. Another important algorithm for explanation is the \textit{Permutation Importance}\cite{10.1093/bioinformatics/btq134}, which we used in our experiments through the ELI5 package to compute the importance of features for prediction.



\section{Experimental Setup and Methodology}
In this section, the methodology of the experiments is described including the dataset, preprocessing and classification methods. 
In terms of hardware, all models were trained with NVIDIA TESLA M60 GPU ACCELERATOR.

\subsection{Dataset}
Since the data is provided by the hospitals wards, MIMIC-III dataset\cite{johnson2016mimic} was not suitable for this benchmark, and it might be considered in future works.\par
In order to form an extensive dataset, we collected data from ward patients from six hospitals that have the \textit{Robô Laura} alert system\footnote{Laura Health: https://www.laura-br.com/}. The hospitals are from different geographic locations within Brazil and all of them provide health services through health insurances and also the Brazilian national public health system (Sistema Único de Saúde [SUS]). The demographic study regarding the data can be seen in Table \ref{tab:GP}.  This is a retrospective observational study, which pooled data from 2016 until October 2019. The features selected for the prediction model were vital sign collections (temperature, oxygen saturation, respiratory rate, blood glucose level and blood pressure), biological sex, age, ward, department, and lenght of hospital stay from each patient, and the target outcome was the patient's mortality during hospitalization. We did not assess mental status (AVPU [alert, verbal, response to pain, unresponsive]).

\begin{table}[]
\caption{Demographic Representation within the Dataset used in this Study}
\label{tab:GP}
\begin{center}
\begin{tabular}{@{}lrr@{}}
\toprule
\multicolumn{3}{c}{\textbf{GLOBAL POPULATION N=121,089}} \\ \midrule
 &  &  \\
\textbf{AGE} & \textbf{\#} &  \textbf{\%} \\
0-15: & 14,916 & 12.3 \\
15-17: & 1,775 & 1.4 \\
18-29: & 17,470 & 14.4 \\
30-39: & 18,305 & 15.1 \\
40-49: & 12,771 & 10.5 \\
50-59: & 15,954 & 13.1 \\
60-69: & 18,213 & 15.0 \\
70+: & 21,685 & 17.9 \\
Mean: & 45 & ($\pm$  24.14) \\
 &  &  \\
\textbf{GENDER} & \textbf{\#} &  \textbf{\%} \\
Female: & 72,377 & 59.7 \\
Male: & 48,695 & 40.3 \\
 &  &  \\
\textbf{\begin{tabular}[c]{@{}l@{}}DAYS FROM\\ ENTRANCE\end{tabular}} & \textbf{\#} &  \textbf{\%} \\
0-2: & 71,837 & 59.3 \\
3-5: & 21,607 & 17.8 \\
6-8: & 9,624 & 7.9 \\
9-11: & 4,993 & 4.1 \\
12+: & 12,974 & 10.7 \\
Mean: & 5.20 & ($\pm$  11.33 ) \\
 &  &  \\
\textbf{\begin{tabular}[c]{@{}l@{}}DEATH DURING\\ HOSPITAL STAY\end{tabular}} & \textbf{\#} &  \textbf{\%} \\
YES: & 6,108 & 5.1 \\
NO: & 114,981 & 94.9 \\
 &  &  \\
\textbf{ICU STAY} & \textbf{\#} &  \textbf{\%} \\
YES: & 2,403 & 1.9 \\
NO: & 118,686 & 98.1 \\
&&\\
\textbf{\begin{tabular}[c]{@{}l@{}}ATTENDANCE \\PER HOSPITAL\end{tabular}} & \textbf{\#} &  \textbf{\%} \\
H1: & 16,485 & 13.6 \\
H2: & 28,109 & 23.2 \\
H3: & 17,091 & 14.1 \\
H4: & 19,737 & 16.3 \\
H5: & 6,524 & 5.4 \\
H6: & 33,143 & 27.4 \\

\bottomrule
\end{tabular}
\end{center}
\end{table}


\subsection{Preprocessing}

\begin{figure}
    \centering
    \includegraphics[scale=0.9]{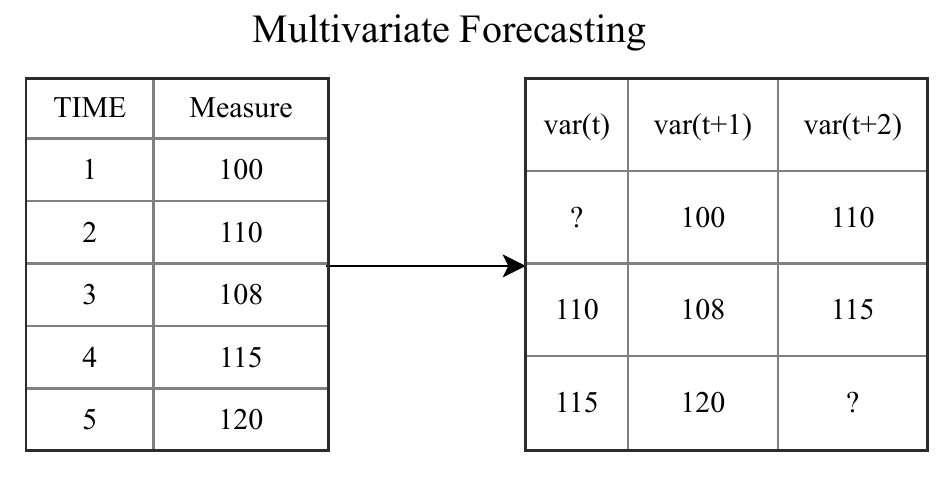}
    \caption{Example of multivariate time series, where the time on the longitudinal table (left) is formatted into an explicit order dependence table (right).}
    \label{fig:forecast}
\end{figure}

Due to the manual imputation of the data at the hospitals, the data is originally longitudinal. Because of this, therein lies the need for preprocessing in order to convert into a time-series format (Figure \ref{fig:forecast}). In order to do this, we must remove outlier values and patients with just one vital sign collected. Missing data is also imputed with the filling-forward process for two timestamps, and after this, we utilize a Random Forest \cite{stekhoven2012missforest} algorithm to predict the remaining missing values, should any be present. This is required since some of the predictive models do not support missing values. For creating the training data, the considered features were a 36 hour time window of vital signs before the patient outcome. Within this time window, the final 12 hours were discarded to prevent model bias. After excluding the 12-hour time frame before the outcome, we were able to use the remaining 24 hours of the 36-hour time window and, during this period, we considered 5 time stamps of the collected vital signs.

\subsection{Classification}
For benchmarking the machine learning models, 
in order to compare the statistical metrics, and because of the imbalance of the data, the best cutoff value (optimal decision threshold) was took into consideration, through cost/consequences of either mistake of 1:10, which means that a FNR (false negative rate) value is 10 times worse than a FPR (false positive rate). This is important since errors are not equal; to classify a dying patient as healthy is much worse than the other way around since it could lead to death. On the other hand, to classify a healthy patient as at risk is an error that leads to increased overhead times for the healthcare team, which is far less of an issue compared to the alternative. \par

 The preliminary statistical classification methods used to validate the data were three: \textit{Leave One Group Out}, \textit{k-fold Cross-Validation}, in this case k=10, and \textit{Windowing Validation}.
\textit{Leave One Group Out} is a cross-validation method, similar to k-fold, in which each fold represents one hospital in this study. Initially it assigns five out of the six groups for training and afterwards it tests on the remaining group, and loops until all the possibilities are done. In this same model, for validation, as the classes were dichotomous, the Kruskal Wallis Test\cite{mckight2010kruskal} was used to find the \textit{p-value} between each feature and the classes for each hospital group. For \textit{p-value} $<$ 0.05 the feature remained significant for the ML model. Given the proportion of features that the \textit{p-value} $>$ 0.05 were 0.02 for H1, 0.01 for H2, 0.02 for H3, 0.04 for H4, 0.05 for H5 and 0.01 for H6. The lacking of non-significative features in the folds of Leave on Group Out had as consequence the slightly different performance in terms of accuracy for each group.
This method then shows generalization between differing hospitals. 
\textit{Windowing Validation} is another method that uses time frames. In this case there can be used ML, MEWS and NEWS2 to analyze the data. ML uses forward-step, timeseries folding, while NEWS2 and MEWS are protocols that use only the last vital sign of the iteration. In our case there were five time frames available in the 24-hour bundle (t-4, t-3, t-2, t-1, and t) (Table \ref{tab:winresults}).
In order to put the results in perspective alongside prior studies, specially in the medical field, the AUC of MEWS and NEWS2 were applied for validation.

\section{Results}





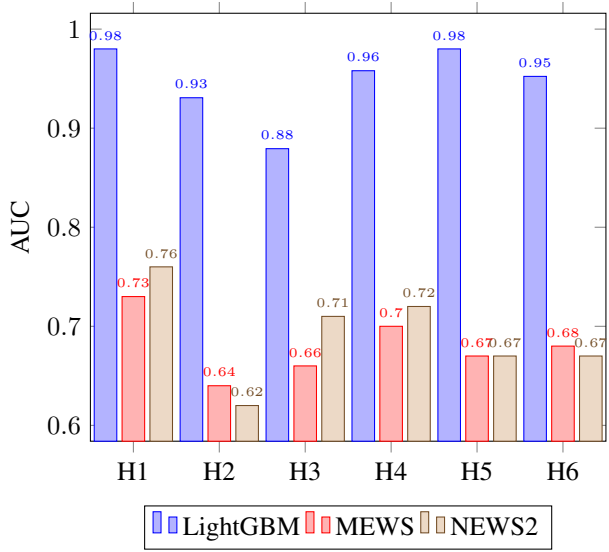
\begin{figure}
\centering
\begin{tikzpicture}[scale=1.0]
\begin{axis}[
    ybar,
    legend style={at={(0.5,-0.15)},
      anchor=north,legend columns=-1},
    ylabel={AUC},
    symbolic x coords={H1, H2, H3, H4, H5, H6},
    xtick=data,
    every node near coord/.append style={font=\tiny},
    bar width=0.3cm,
    nodes near coords,
    nodes near coords align={vertical},
    ]

\addplot coordinates {(H1, .98) (H2, .930721) (H3, .879299) (H4, .957988) (H5, .98) (H6, .952274)};
\addplot coordinates {(H1, .73) (H2, .64) (H3, .66) (H4, .70) (H5,0.67) (H6, .68)};
\addplot coordinates {(H1, .76) (H2, .62) (H3, .71) (H4, .72) (H5,0.67) (H6, .67)};

\legend{LightGBM, MEWS, NEWS2}
\end{axis}
\end{tikzpicture}
\caption{(LightGBM x MEWS x NEWS2) per Hospital}
\label{fig:benchperhosp}
\end{figure}

The results were compiled in Table \ref{tab:mainresults}, Table \ref{tab:winresults} and Figure \ref{fig:benchperhosp}. It can be observed that among six statistical classification methods used, the \textit{LightGBM, XGBoost} and \textit{CatBoost}, denoted the best evaluation parameters, via folded cross-validation, Leave One Group Out and Windowing Validation, and this surpasses the MEWS and NEWS2 protocols by around 20 percentage points. In Laura system, LightGBM model is used in production for real-time clinical deterioration prediction, and it should perform better than the others gradients models because on its implementations is used the novel technique \textit{Gradient-based One-Side Sampling (GOSS)} that keeps all the instances with large gradients and performs random sampling on the instances with small gradients to estimate the information gain \cite{NIPS2017_6907}.

\begin{table}[hbt!]
\caption{AUC and F1 of algorithms for 10CV and Leave One Group Out classification}
\label{tab:mainresults}
\begin{tabular}{@{}lcccc@{}}
\toprule
\multicolumn{1}{c}{\multirow{2}{*}{\textbf{Algorithm}}} & \multicolumn{2}{c}{\textbf{10-fold Cross Validation}} & \multicolumn{2}{c}{\textbf{Leave One Group Out}} \\ \cmidrule(l){2-5} 
\multicolumn{1}{c}{} & \textbf{AUC} & \textbf{F1} & \textbf{AUC} & \textbf{F1} \\ \midrule
LightGBM        & \bf{.961} & \bf{.671} & \bf{.949} & .620 \\
XGBoost         & .956 & .632 & \bf{.947} & \bf{.649} \\
CatBoost        & .955 & .643 & .935 & .559 \\
Random Forest   & .940 & .609 & .933 & .584 \\
Log. Regression & .932 & .556 & .932 & .573 \\
Naive Bayes     & .841 & .379 & .853 & .363 \\
\midrule
NEWS2           & .704 & .196 & .704 & .196 \\ 
MEWS            & .697 & .175 & .697 & .175 \\
\bottomrule
\end{tabular}
\end{table}

\subsection{Explainable Alerts}
\begin{figure}[ht]
    Alert: Updated
    \begin{itemize}
      \item Heart Rate: \textcolor{red}{\bf{204}}
      \item Systolic Blood Pressure: \textcolor{red}{\bf{47}}
      \item Calcium: \textcolor{red}{\bf{1.34}}
    \end{itemize}
    \caption{Example of \textit{Robô Laura } platform explaining why the alert was generated.}
    \label{fig:timeline2}
\end{figure}

The \textit{Robô Laura} monitoring system is deployed with ELI5 to produce visual explanation in order to overcome the eventual shortcomings described previously\cite{thurier2019inspecting}. (Fig. \ref{fig:timeline2})

\subsection{Windowing Validation}

Vital signs are obtained on average of every six hours. As seen in Table \ref{tab:winresults} the best result considering the Area Under the ROC Curve is using all five data points for all algorithms. In every experiment, the ML models benefited from previous data points to make its decision. MEWS also got its best performance with the last vital signs observations. The best result for MEWS was using threshold '\textgreater1' and NEWS2 was using threshold '\textgreater2'.

\begin{table}[hbt!]
\caption{Windowing results validation (AUC), is evaluated from 12h to 36h before patient outcome (this results in one to five data points), where 12h is noted as  \textit{t} and 36h as \textit{t-4}.}
\label{tab:winresults}
\begin{center}
\begin{tabular}{@{}lrlrrr@{}}
\toprule
\textbf{Algorithm} & \textbf{t-4} & \textbf{t-3} & \textbf{t-2} & \textbf{t-1} & \textbf{t} \\ \midrule
LightGBM        & .935 & .943 & .949 & .956 & \bf{.961} \\
XGBoost         & .928 & .937 & .944 & .950 & \bf{.956} \\
CatBoost        & .930 & .938 & .944 & .950 & \bf{.955} \\
Random Forest   & .906 & .914 & .923 & .933 & \bf{.940} \\
Log. Regression & .905 & .914 & .920 & .928 & \bf{.932} \\
Naive Bayes     & .858 & .833 & .831 & .836 & \bf{.841} \\
\midrule
NEWS2           & .645 & .651 & .659 & .678 & \bf{.705} \\
MEWS            & .658 & .674 & .677 & .689 & \bf{.697} \\ 
\bottomrule
\end{tabular}
\end{center}
\end{table}

This result shows that the relation between several vital signs gives the models more features to understand clinical deterioration. For such reason, we used, in previous sections, all five data points to perform the experiments.

\section{Future Work and Conclusion}

Brazilian hospitals have EHR's with no standardization. Moreover, laboratory examinations are not frequent enough for early warning risk identification and vital signs, and more frequently, they do not perform real-time imputation. With that, it is necessary to have a robust system that can have a great impact on early identification of patients at risk and, if possible, that can take into account the patient's history, with data that is often available. Even the classic Machine Learning algorithms we have shown to have a great capacity of being used in hospitals of different capacities, and also have shown to be better than traditional EWS protocols. \par

With the hypothesis of the alternative or the supplement AI system for predicting clinical deterioration in hospital wards, the algorithm that showed the best results was \textit{LightGBM} with AUC of 0.961 and F1 of 0.671, which far exceeds the respective scores of 0.697 and 0.175 by the MEWS system. Further work will be able to explore the warn workflow retrospectively and compare the changes due to the AI early warning systems, improving the model explanation to guarantee the communication with the assistant team. \par

Additionally, we will explore more complex algorithms such as Deep Learning models in order to obtain a more accurate state-of-the-art benchmark, through an extensive evolutionary hyperparameter search in order to customize a deep neural network for patient outcome classification\cite{bird2019evolutionary,bird2019deep,miikkulainen2019evolving}. We will also consider a larger amount of patient historical data and aim to generalize the algorithms for any integrated hospital by transfer learning results, as was seen within this study. \par

The sample dataset
is available at the project's GitHub Page\footnote{https://github.com/laura-health/cbms2020}, in order to be easily replicated. The sample dataset has no patient identification data due to privacy reasons; it contains only a subset of the original dataset. 

\section*{Competing Interests}
All members of Laura Health work towards the development of \textit{Robô Laura}, which is a commercial software alert system to identify patient clinical deterioration. All other authors have no other competing interests to declare.

\bibliographystyle{ieeetr} 
\bibliography{bibliography} 
\vspace{12pt}
\end{document}